# ECG-Lens: Benchmarking ML & DL Models on PTB-XL Dataset

Saloni Garg
The LNM Institute of Information
Technology, Jaipur, India

Ukant Jadia
Sir Padampat Singhania University
Udaipur, Rajasthan, India

Amit Sagtani
The LNM Institute of Information
Technology, Jaipur, India

Kamal Kant Hiran
Sir Padampat Singhania University
Udaipur, Rajasthan, India

*Abstract*— Automated classification of electrocardiogram (ECG) signals is a useful tool for diagnosing and monitoring cardiovascular diseases. This study compares three traditional machine learning algorithms (Decision Tree Classifier, Random Forest Classifier, and Logistic Regression) and three deep learning models (Simple Convolutional Neural Network (CNN), Long Short-Term Memory (LSTM), and Complex CNN (ECG-Lens)) for the classification of ECG signals from the PTB-XL dataset, which contains 12-lead recordings from normal patients and patients with various cardiac conditions. The DL models were trained on raw ECG signals, allowing them to automatically extract discriminative features. Data augmentation using the Stationary Wavelet Transform (SWT) was applied to enhance model performance, increase the diversity of training samples, and preserve the essential characteristics of the ECG signals. The models were evaluated using multiple metrics, including accuracy, precision, recall, F1-score, and ROC-AUC. The ECG-Lens model achieved the highest performance, with 80% classification accuracy and a 90% ROC-AUC. These findings demonstrate that deep learning architectures, particularly complex CNNs substantially outperform traditional ML methods on raw 12-lead ECG data, and provide a practical benchmark for selecting automated ECG classification models and identifying directions for condition-specific model development.

## I. Introduction

Cardiovascular diseases pose a major public health challenge worldwide, causing significant harm to people's lives and healthcare systems. Despite the huge progress in medical science and technology, heart-related illnesses are still the primary cause of mortality on a global scale, resulting in a concerning number of deaths each year.[1]. The impact of cardiovascular disorders cannot be overstated, as they can cause significant damage to the heart, blood vessels, and organs, leading to long-term and short-term health consequences and reducing the quality of life for those affected[2]. That's why it's crucial to detect and diagnose these diseases early, as this holds the key to mitigating their devastating impacts and improving patient outcomes through timely and appropriate interventions.

The electrocardiogram (ECG) is a cardio-vascular diagnostic and monitoring tool that generates a graphical representation of the electrical activity of the heart. It's an indispensable tool in the field of cardiology, which enables healthcare professionals to monitor and evaluate the functionality of the heart, detect potential abnormalities, and guide targeted treatment strategies. The ECG records the electrical impulses that control the heart's rhythmic contractions, which are captured in intricate patterns that can be analysed to derive insights into the heart's functioning.[3] However, analysing ECG readings traditional way requires a great deal of time and expertise, and it can be prone to human error and subjectivity. This is particularly problematic in regions where access to skilled medical personnel is limited, which can result in delayed diagnoses and inadequate treatment.

The field of machine learning and artificial intelligence has progressed rapidly in the last decade, and recent developments have resulted in a significant shift in how we approach cardiovascular disease and analyze electrocardiograms (ECGs). These advancements offer a tantalizing prospect of automating ECG analysis with unprecedented accuracy, efficiency, and accessibility[4]. To unleash the wealth of information contained in ECG data, multiple solutions have been proposed, including demographic analysis[5], wearable device data analysis[6], and sophisticated signal processing techniques[7]. However, while these approaches have shown some promising results, they often lack the level of detail and also specificity required for precise diagnosis and comprehensive risk stratification across the full spectrum of cardiac conditions.

Accurately interpreting electrocardiograms (ECGs) is essential for identifying a wide range of cardiac abnormalities, including arrhythmias, ischemia, and structural defects[8]. This critical function plays a vital role in creating targeted and effective treatment plans tailored to meet the unique needs of each patient. The ability to distinguish between these subtle nuances is often the determining factor between life and death, underscoring the urgent need for reliable and robust automated systems to classify ECGs.

The focus of this current research is to conduct a comprehensive comparative analysis of traditional machine learning models and advanced deep learning architectures for effectively classifying ECG statements into five distinct categories(superclasses) using the well-known PTB-XL(Physikalisch-Technische Bundesanstalt - Extended Long - Term) ECG dataset[9]. This study aims to do the evaluation and benchmark the performance of these models across a diverse range of cardiac abnormalities, with the goal of identifying the most accurate and reliable approach for automated ECG interpretation.

This paper comprises five sections: a literature review in Section 2, methodology in Section 3, results analysis in Section 4, and conclusions in Section 5. In our study, we have utilized various machine learning algorithms and deep learning architectures. The primary objective of this study is to enhance the field of automated ECG analysis, which will ultimately enhance cardiac care for patients worldwide.

TABLE I. ECG CLASSES, COUNTS, DESCRIPTIONS, AND MEDICAL TERMS IN THE PTB-XL DATASET

| Notation | Class | Description | Medical Term | Records |
|---|---|---|---|---|
| 0 | NORM | Normal ECG pattern | Sinus Rhythm | 7293 |
| 1 | MI | Myocardial Infarction | Heart Attack | 4103 |
| 2 | ST/T Change | ST-segment or T-wave changes | Ischemia, Injury | 3869 |
| 3 | CD | Conduction Disturbance | Bundle Branch Block | 3790 |
| 4 | HY | Ventricular Hypertrophy | Enlarged Ventricle | 2782 |

## II. LITERATURE REVIEW

Accurately classifying 12-lead electrocardiogram (ECG) signals is crucial for medical data analysis and has been the focus of extensive research. Many studies have aimed to create effective models for this purpose. The following provides a comprehensive overview of relevant research in this field.

A study by Acharya et al. [10] suggests a deep learning-based method for diagnosing cardiac arrhythmias using ECG signals. The researchers employed a 19-layer convolutional neural network (CNN) model to categorize ECG recordings for different types of cardiac arrhythmias condition. The study highlights the potential of deep learning techniques in ECG signal classification, as the model achieved an accuracy of 93.5% on a publicly available dataset.

In their research, Yildirim et al. [11] proposed a hybrid model that combines wavelet transform and support vector machines (SVM) to classify ECG signals using the MIT-BIH dataset. The model extracts relevant features from ECG signals through wavelet transform and uses them with an SVM classifier to classify them. The model achieved an impressive accuracy of 99.5% on a publicly available dataset, demonstrating the potential of combining feature engineering and machine learning techniques for ECG classification.

Smigiel et al.[12] conducted a study where they assessed the performance of convolutional neural networks (CNNs), SincNet, and a CNN with entropy-based features on the PTB-XL dataset for multi-class ECG classification. The CNN achieved an accuracy of 88.2% for 2 classes and 72.0% for 5 classes. SincNet achieved an accuracy of 85.8% for 2 classes and 73.0% for 5 classes. However, their novel CNN, which had entropy features, achieved the highest accuracy of 89.2% for 2 classes, 76.5% for 5 classes, and 69.8% for 20 classes. Although SincNet performed slightly better than the basic CNN on more detailed classification tasks, the entropy-augmented CNN was superior across all tasks evaluated on this extensive clinical ECG dataset.

Jambukia et al. [13] presented a comprehensive survey of ECG classification into arrhythmia types. The paper discusses the issues involved in ECG classification and also provides a detailed review of preprocessing techniques, ECG databases, feature extraction methods, artificial neural network-based classifiers, and performance measures. The authors conclude that neural networks are generally effective and preferable for ECG classification, with multilayer perceptron neural networks (MLPNN) providing good accuracy for ECG beat classification.

Luz et al. [14] conducted a comprehensive survey on ECG-based heartbeat classification for arrhythmia detection using multiple datasets such as the MIT-BIH Arrhythmia Database, the St. Petersburg INCART 12-lead Arrhythmia Database, and the CUDB Database. This paper provides a detailed review of various machine learning techniques, which are support vector machines (SVMs), neural networks, decision trees, and ensemble methods. The authors discuss the challenges involved in ECG signal analysis, such as noise, baseline wander, and morphological variability.

In their study, Rajpurkar et al. [15] reviewed the utility of convolutional neural networks (CNNs) for arrhythmia detection at a cardiologist level. They trained the CNNs on a large dataset of over 64,000 single lead ECG recordings, collected from the PhysioNet Challenge 2017 database. The study focuses on the application of deep learning techniques for ECG analysis. The authors explain the benefits of CNNs compared to traditional machine learning methods, such as their ability to automatically learn relevant features from raw data. They also emphasize the importance of large-scale ECG datasets for training robust deep-learning models.

Furthermore, while some studies have utilized datasets like PTB-XL, which comprises 21,837 clinical standard 12-lead ECG recordings taken from 18,885 patients, more expansive and diverse datasets are necessary to capture the full scope of variations present in real-world clinical settings. Therefore, this project will also consider the use of larger and more diverse datasets to ensure that the model can capture the full range of ECG abnormalities encountered in clinical practice.

## III. METHODOLOGY

### A. Data Preprocessing

The PTB-XL dataset used in this project contains 21,837 clinical ECG recordings from 18,885 unique patients[9]. These recordings are categorized into 5 superclasses and 24 subclasses. We have taken the most relevant and prevalent classes for this study, which are Normal ECG, Myocardial Infarction, ST/T Change, Conduction Disturbance, and Hypertrophy. Table I displays the different classes along with their descriptions. It also includes the notation used in the project to better understand the preprocessing workflow. Additionally, the table contains a brief description of the various medical terms used for each type of ECG class and the total number of samples of each class that are present in the PTB-XL dataset.

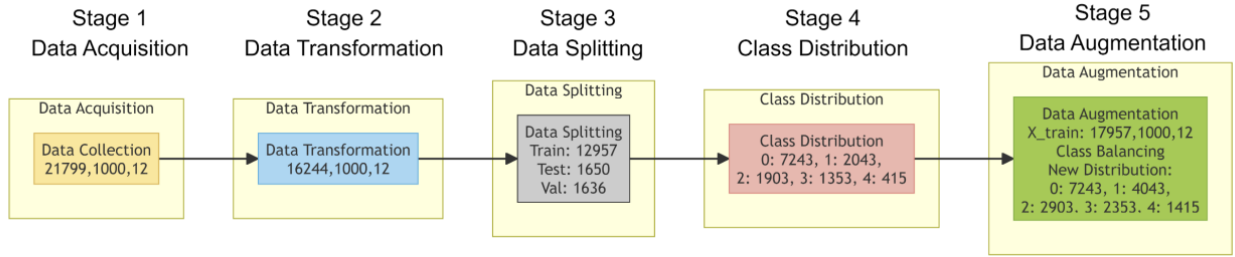

Fig. 1. Data Preprocessing Pipeline for ECG Dataset

The data acquisition and preprocessing workflow is described as follows: The PTB-XL dataset, comprising 21,799 standard 12-lead ECG recordings of 10 seconds duration at a sampling frequency of 100 Hz, was acquired [3]. Each ECG recording consists of 12 channels (leads), with each channel containing 1,000 data points. Therefore, each ECG sample has a shape of (1,000, 12), where 1,000 represents the number of data points per lead, and 12 represents the number of leads. The project retained five relevant superclasses, resulting in a reduction of the dataset to 16,243 instances, each with a shape of (1,000, 12). The data was then split into train (12,957 instances), test (1,650 instances), and validation (1,636 instances) sets, preserving the original shape of (1,000, 12) for each instance. The training set exhibited class imbalance, which was mitigated through oversampling minority classes (MI, ST/T, CD, and Hy) by adding 1,000 samples to each class using the Stationary Wavelet Transform (SWT) [16] data augmentation technique. This resulted in a balanced training set with 17,957 instances, each with a shape of (1,000, 12). For the traditional machine learning models, the 3D data (n, 1000, 12) was reshaped to 2D (n, 12000) by flattening the 12 leads into a single feature vector where n is the sample count. Conversely, for the deep learning models, the target variables underwent one-hot encoding, and the input data was normalized to facilitate efficient training and convergence. The data preprocessing pipeline, which includes data acquisition, transformation, splitting, class distribution, and augmentation stages, is visually shown in Fig. 1. This comprehensive approach ensured that the data was prepared correctly and optimized for both machine learning and deep learning models, enabling a fair and robust comparative evaluation of their performance for classifying ECG signals.

Electrocardiography (ECG) is a medical monitoring technique that measures the electrical activity of the heart using a standard 12-lead. It is commonly used as a diagnostic tool in cardiology that helps to identify a range of cardiac conditions. To help with a better understanding, here is a detailed description of the five superclasses related to ECG:

The Normal ECG class shows a healthy cardiac rhythm and electrical conduction. It displays characteristics of P waves, QRS complexes, and T waves all within normal ranges, which serve as a baseline for comparison against abnormal ECG patterns.

The Myocardial Infarction class is of critical importance since it indicates a heart attack, which is a medical emergency. ECG patterns in this class may show ST-segment elevations or depressions, abnormal Q waves, or T-wave inversions, depending on the location and extent of the myocardial damage.

The class ST/T Change covers various ECG abnormalities that are connected to the ST segment and T wave. These abnormalities indicate myocardial ischemia, disruptions in electrolyte balance, or the influence of particular medications. The changes may involve ST-segment elevation or depression, T-wave inversions, or unusual T-wave shapes.

Conduction Disturbance shows ECG patterns that indicate disturbances in the electrical conduction system of the heart. These disturbances can appear as prolonged PR intervals, also known as atrioventricular blocks, widened QRS complexes, which are referred to as bundle branch blocks, or other conduction delays or blockages that can result in abnormal heart rhythms or arrhythmias.

The Hypertrophy class comprises the ECG patterns associated with an enlarged or thickened heart muscle, often seen with conditions like hypertension and some heart valve diseases. These patterns include increased QRS voltage, ST-segment changes, and abnormal T-wave morphologies, reflecting the change in electrical activity of the hypertrophied heart muscle.

As classes are imbalanced as shown in Records of Table I, to address this and improve model performance, augmentation is performed on the minority classes. We used the Stationary Wavelet Transform (SWT)[16] algorithm for augmentation, as it overcomes the limitations of other methods and preserves signal characteristics. This algorithm applies augmentation techniques to the wavelet coefficients in different frequency bands and then reconstructs the augmented signal[16].

*B. Implementation*

The implementation of this study follows a comprehensive pipeline, as illustrated in the project flow chart Fig. 2

, beginning with data preprocessing, model implementation, and comparative evaluation. The first stage involves careful data preprocessing, including data collection, validation, transformation, and augmentation using the Stationary Wavelet Transform (SWT) method. This stage of the pipeline ensures that the ECG data is properly prepared and enhanced for effective model training and analysis.

After the data preprocessing stage, the project flow diverges into two parallel paths of implementation: machine learning (ML) models and deep learning (DL) models. The ML models used in this comparison analysis are Logistic Regression, Decision Tree, and Random Forest Classifier, which are well-established techniques in the field of ECG analysis. On the other hand, the DL models include Convolutional Neural Networks (CNN), Long Short-Term Memory (LSTM) networks, and a specialized model called ECG-Lens, which is designed to capture the intricacies of ECG signals.

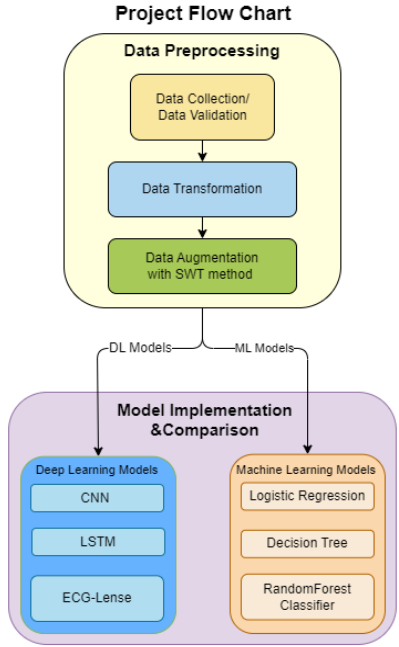

Fig. 2. *Flow chart outlining the stages of data preprocessing, model implementation, and result comparison.*

TABLE II: COMPARATIVE PERFORMANCE EVALUATION OF TRADITIONAL MACHINE LEARNING MODELS

| Metrics | Random Forest Classifier(%) | Decision Tree(%) | Logistic Regression(%) |
|---|---|---|---|
| Accuracy | 54 | 52 | 40 |
| ROC-AUC | 62 | 58 | 50 |
| F1-Score | 23 | 22 | 21 |
| Precision | 25 | 24 | 21 |
| Recall | 24 | 24 | 22 |

These diverse models are included in the "Model Implementation & Comparison" stage, where their performance and behavior are carefully evaluated using metrics, like accuracy, precision, recall, F1-score, and ROC-AUC. This comparative analysis is done to highlight the strengths and limitations of each approach, ultimately determining the most effective model for ECG-based heart disease classification.

*1) Machine Learning Implementation*

*a) Decision Tree*

Decision Trees can be used to recursively partition the feature space in order to make accurate classifications, and they are particularly effective for identifying intricate non-linear patterns in ECG data [17]. The fundamental idea behind Decision Trees involves dividing the feature space based on the feature that offers the most valuable information gain, resulting in a tree-like structure of decision rules. However, Decision Trees are prone to overfitting, which can be addressed through methods such as pruning and ensemble techniques.

The results show that the decision tree model performed reasonably well, resulting in a moderate accuracy score of 52%. This indicates that it successfully identified some of the underlying patterns in the ECG data. However, the relatively low F1 score of 22% and ROC-AUC score of 58% suggest that the model may have encountered some difficulties in accurately classifying all the classes, particularly the less frequent ones due to overfitting.

*b) Logistic Regression*

Logistic Regression is widely used as a linear classification algorithm that models the likelihood of an outcome. The basic idea behind Logistic Regression is to model the likelihood of an outcome (in this case, ECG signal classification) using a logistic function. While this approach may have some limitations in capturing the complex non-linearities present in the ECG data, it provides a dependable and understandable baseline for ECG classification.

In this study, the logistic regression model had an accuracy of 40%, as well as a lower-than-expected F1 score of 21% and ROC-AUC score of 50%. These findings highlight the linear nature of the logistic regression model may have constrained its ability to capture the intricate non-linear patterns and relations inherent in the ECG signals.

*c) Random Forest Classifier*

Random Forest is a machine learning method that combines multiple decision trees to improve the accuracy and robustness of classification. It is specifically useful for ECG data analysis, as it can handle the complex non-linear relationships present in ECG signals [18].

The Random Forest classifier works by utilizing the predictive strength of multiple decision trees. This approach helps to avoid overfitting problems that can arise when using single decision trees. By utilizing a group of decision trees on random subsets of data points and features, the Random Forest model is able to capture intricate patterns and relations in ECG data more effectively.

The random forest classifier exhibits the strongest performance, with an accuracy of 54%, an F1 score of 23%, and an ROC-AUC score of 62%. These results align with the reasoning that the ensemble technique of the random forest model, which uses the power of multiple decision trees collectively for prediction, can effectively handle the non-linearities present in the ECG data.

A comprehensive comparison of three traditional machine learning models, namely Decision Tree Classifier, Logistic Regression, and Random Forest Classifier, was evaluated on an ECG classification task, in Table II.

Traditional machine learning models raise challenges in ECG classification[21]. Decision trees result the moderate accuracy but lower F1-scores and ROC-AUC due to overfitting and sensitivity to irrelevant features. Logistic regression struggles with complex non-linear ECG data patterns. On the other hand, random forest classifiers outperform both approaches by handling non-linearities, and high dimensionality to a certain level. These findings underscore the importance of carefully selecting machine learning models that are able to capture the intricate relationships and unique characteristics of ECG signals. Random forest classifiers illustrate the potential advantages of ensemble techniques. Ultimately, traditional models provide a valuable baseline for comparison with advanced deep-learning approaches.

*2) Deep Learning Implementation*

Convolutional Neural Networks (CNNs) are a type of deep learning model that is particularly well-suited for processing and analyzing large data. The key components of a CNN are:

*Convolutional Layers (Conv1D)*: A series of trainable filters are applied to the input data by these layers, which enables the model to extract important characteristics from the input.

*Pooling Layers (MaxPooling1D)*: The following layers decrease the resolution of the feature maps, diminishing the spatial dimensions and extracting the crucial features.

*Dense Layers (Dense):* Dense layers, also referred to as fully connected layers, utilize the extracted features to perform predictions or classifications.

*Flatten Layer (Flatten):* The Flatten layer is used to convert the 2D feature maps into a 1D vector, this layer reshapes the data to be given as input into the Dense layers.

*Dropout Layer (Dropout):* Dropout, as a regularization method, involves randomly deactivating a portion of the neurons while training to prevent the model from overfitting.

*Global Average Pooling (GlobalAveragePooling1D):* This layer conducts a global average pooling operation on the feature maps, to decrease the spatial dimensions to a single value per feature map.

*Batch Normalization (BatchNormalization):* The inputs to each layer are normalized by this layer, which can improve training stability and model performance.

*Activation Functions (ReLU, Activation):* These activation functions are used to introduce non-linearity into the model, enabling it to confidently learn complex patterns in the data with the best accuracy and precision.

*Long Short-Term Memory (LSTM):* LSTM, a form of recurrent neural network (RNN), is especially adept at handling sequential data, such as time series or text.

In this section, we investigate three deep learning models for the classification of raw 12-lead electrocardiogram (ECG) data: a Convolutional Neural Network (CNN) model, a Long Short-Term Memory (LSTM) model, and a high-performance ECG-Lens model. Each of these models was designed and trained to classify the ECG data into five distinct classes.

*a) Simple CNN*

The initial model is a CNN-based architecture, featuring a series of 1D convolutional layers with an input shape of (n,1000,12) here n is the number of samples, max-pooling layers, a flattening layer, dense layers, and a dropout layer. We compiled this model with the Adam optimizer, categorical cross-entropy loss, and a range of evaluation metrics, including Precision, Recall, Receiver Operating Characteristic - Area Under the Curve(ROC-AUC), Categorical Accuracy, and Accuracy. The model achieved an accuracy rate of 71%, an ROC-AUC of 85%, and an F1-score of 69%. However, the confusion matrix revealed significant misclassifications across the various classes.

TABLE III: COMPARATIVE PERFORMANCE EVALUATION OF ADVANCED DEEP LEARNING MODELS

| Metrics | ECG-Lens Model(%) | LSTM Model(%) | CNN Model(%) |
|---|---|---|---|
| Accuracy | 80 | 73 | 71 |
| ROC-AUC | 90 | 87 | 85 |
| F1-Score | 78 | 72 | 69 |
| Precision | 80 | 78 | 73 |
| Recall | 76 | 71 | 66 |

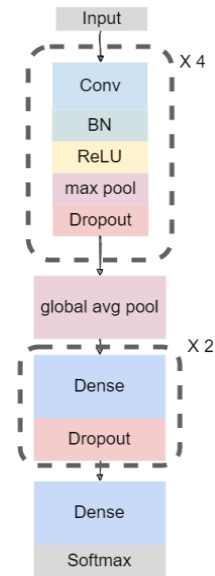

Fig. 3. *The neural network architecture, it contains 4 layers of convolution and 3 fully-connected.*

*b) LSTM*

The second model employs an LSTM architecture and uses multiple LSTM layers, as well as batch normalization and dropout layers, all layers followed by dense layers. Then the model is compiled with the Adam optimizer, categorical cross-entropy loss, and the identical evaluation metrics as the CNN model. During testing, this model achieved an accuracy of 73%, ROC-AUC of 87%, and F1-score of 72%. The confusion matrix for the LSTM model displayed a reasonably even performance across the classes.

*c) ECG-Lens Model*

The high-performance ECG-Lens model, which is the third and best-performing model in this study, is a CNN model especially focused on classifying the raw ECG data. It boasts a complex architecture that is customized to capture the intricate patterns and nuances present in ECG signals. As shown in Fig. 3, the ECG-Lens model comprises multiple 1D convolutional layers with an increasing number of filters, ranging from 128 to 1024. These convolutional layers are responsible for extracting and learning discriminative features from the raw ECG data. For the prevention of overfitting and to improve generalization, batch normalization layers are incorporated after each convolutional layer, followed by ReLU activations and max-pooling layers. The model also employs a high dropout rate of 0.4 to further mitigate the risk of overfitting.

## Confusion Matrix

|  | CD | HYP | MI | NORM | STTC | |
|---|---|---|---|---|---|---|
| **CD** | 871 / 52.79% | 95 / 5.76% | 61 / 3.70% | 60 / 3.64% | 28 / 1.70% | 1115 |
| **HYP** | 12 / 0.73% | 140 / 8.48% | 17 / 1.03% | 8 / 0.48% | 1 / 0.06% | 178 |
| **MI** | 16 / 0.97% | 16 / 0.97% | 160 / 9.70% | 7 / 0.42% | 2 / 0.12% | 201 |
| **NORM** | 2 / 0.12% | 3 / 0.18% | 0 / 0.00% | 106 / 6.42% | 1 / 0.06% | 112 |
| **STTC** | 11 / 0.67% | 2 / 0.12% | 4 / 0.24% | 3 / 0.18% | 24 / 1.45% | 44 |
|  | 912 | 256 | 242 | 184 | 56 | 1650 |

(Predicted vs Actual)

Fig. 4. *Confusion matrix evaluating the classification performance of the ECG-Lens model across five classes, with the final row and column showing the total number of actual and predicted instances.*

After using the convolutional and pooling layers, the ECG-Lens model adds a global average pooling layer that consolidates the spatial information throughout the entire signal. Then two dense layers were added that had a dropout of 0.4. The last dense layer uses a softmax activation function to calculate the probabilities for the five targeted superclasses. The architecture of the ECG-Lens model is depicted in Figure 3.

During the training process, the ECG-Lens model was compiled with the Adam optimizer with a learning rate of 0.001 and categorical cross-entropy loss function. To evaluate the model's effectiveness, various performance metrics, including precision, recall, area under the receiver operating characteristic curve (ROC-AUC), categorical accuracy, and F1-score, were used.

The ECG-Lens model demonstrated outstanding performance, achieving an impressive accuracy rate of 80%, an ROC-AUC score of 90%, and an F1 score of 78%. The confusion matrix for this model further spotlights its ability to accurately classify ECG signals across the five super-classes, outperforming the other models implemented in this study.

The comparative analysis of the deep learning models showcased the effectiveness of the high-performance ECG-Lens architecture, Model 3, in accurately classifying the 12-lead ECG data. The model's deeper and more complex structure, along with the use of advanced techniques like batch normalization and increased dropout, enabled it to outperform the other models with metrics like accuracy, ROC-AUC, and F1 score. These findings suggest that the ECG-Lens model could be a valuable tool for heart disease classification using raw ECG data.

Table III consists of three advanced deep learning models – ECG-Lens, LSTM, and CNN – that are compared in their ability to classify heart diseases using ECG(electrocardiogram) signals. The results reveal that the ECG-Lens model performs best in comparison to the other models across all evaluation metrics, including Accuracy (80%), ROC-AUC (90%), F1-Score (78%), Precision (80%), and Recall (76%). The LSTM model slightly follows, surpassing the CNN model in most metrics. This comparative analysis underscores the effectiveness of these deep learning techniques, and especially the ECG-Lens architecture, in providing precise diagnoses of heart conditions based on ECG data.

*3) Performance Matrices*

The performance metrics used in this study consist of a comprehensive set of measures, including accuracy, area under the curve (ROC-AUC), F1-score, precision, and recall. These metrics are highly utilized and recognized in the domain of machine learning and deep learning, enabling a thorough evaluation of the model's performance from multiple perspectives[20].

Accuracy is an important metric that determines the overall accuracy of a model's predictions, providing a broad view of its classification capabilities. However, in situations where class imbalance is present, as is commonly seen with ECG data[19], relying completely on accuracy can be misleading. As a result, supplementary metrics, such as ROC-AUC, F1-score, precision, and recall, are utilized to gain a more detailed understanding of the model's performance. Mathematically accuracy can be presented as

$$\text{Accuracy} = (TP + TN) / (TP + TN + FP + FN) \quad (1)$$

The ROC-AUC can be calculated from the ROC curve, which charts the true positive rate against the false positive rate for multiple threshold values. The ROC-AUC provides a diverse assessment of a model's discriminative way of measuring the balance between true positive and false positive rates.

On the contrary, precision and recall provide valuable information in classification models, highlighting their strengths and weaknesses. Precision examines the ratio of true positive instances among positive predictions, whereas recall assesses the model's capacity to identify all relevant positive instances. The formula of Precision and Recall is

Precision = TP / (TP + FP)     (2)

Recall = TP / (TP + FN)     (3)

The F1 score is a metric that combines recall and precision into a single measure, providing a balanced evaluation of a model's performance. It ranges from 0 to 1. Higher F1 scores indicate better performance. Mathematically

f1-score = 2 × (Precision × Recall) / (Precision + Recall)     (4)

By employing this comprehensive set of performance metrics, this study aims to conduct a thorough evaluation of the traditional algorithms(machine learning) and advanced algorithm(deep learning) capabilities in classifying ECG signals, ultimately contributing to the development of reliable and accurate ECG analysis tools for heart disease management.

## IV. RESULTS

The performance of the various machine learning and deep learning models evaluated in this study is summarized in Tables II and III.

In traditional machine learning models, the Random Forest Classifier demonstrated the strongest performance, achieving an accuracy of 54%, F1-score of 23%, and ROC-AUC of 62% (Table II). The ensembling technique of the Random Forest model, allowed it to effectively handle the non-linear patterns present in the ECG data, outperforming the Decision Tree (accuracy 52%, F1-score 22%, ROC-AUC 58%) and Logistic Regression (accuracy 40%, F1-score 21%, ROC-AUC 50%) models.

The deep learning models demonstrated better performance than the traditional machine learning approaches (Table III). The ECG-Lens model, with its deeper and more complex architecture, achieved the best results, attaining an accuracy rate of 80%, a ROC-AUC of 90%, and an F1-score of 78%. The model's advanced techniques, such as increased filter sizes in the convolutional layers, batch normalization, and a high dropout rate, enabled it to effectively capture the intricate patterns in the ECG data, outperforming the LSTM model (accuracy 73%, ROC-AUC 87%, F1-score 72%) and the simple CNN model (accuracy 71%, ROC-AUC 85%, F1-score 69%).

The ECG-Lens model has shown remarkable progress in comparison to previous research conducted by Smigiel et al. [12]. While the model resulted in a classification accuracy of 76% for 5 superclasses, the ECG-Lens model has demonstrated a significant improvement, with an accuracy rate of 80%. Handling of the class imbalance and model architecture is performed well on the raw ECG data. The higher accuracy achieved by ECG-Lens is especially promising as it may lead to more precise and dependable analysis of ECG signals, ultimately improving the diagnosis and monitoring of heart disease and cardiovascular conditions in clinical practice.

The classification performance of the ECG-Lens model proposed in this study was evaluated using a confusion matrix, which is depicted in Fig. 4. This matrix provides a detailed overview of the model's classification accuracy across five ECG classes: Conduction Disturbance (CD), Hypertrophy (HYP), Myocardial Infarction (MI), Normal (NORM), and ST/T Change (STTC). The matrix's rows correspond to the real classifications, whereas the columns signify the model's predicted classifications. The diagonal elements, highlighted in green, represent the instances that were correctly classified. For instance, out of 1115 instances of the CD class, the model correctly classified 871 (78.03%), with the remaining instances being misclassified into other classes. The model also achieved an impressive accuracy of 83.03% for the NORM class, correctly classifying 900 out of 1084 instances. The last row and column of the confusion matrix provide the total number of instances for each class, enabling a comprehensive understanding of the model's performance and the distribution of instances across classes.

The ECG-Lens model outperformed other models evaluated in this study, such as Long Short-Term Memory (LSTM) and Convolutional Neural Network (CNN), as evidenced by its high F1-score, precision, recall, and accuracy values. This demonstrates the effectiveness of the proposed deep learning architecture in accurately classifying ECG signals into various cardiac abnormalities, which could lead to improved diagnostic capabilities and patient care.

## V. CONCLUSION

This study aimed to evaluate and compare the efficacy of traditional machine learning algorithms and advanced deep learning models for the classification of electrocardiogram (ECG) signals. The outcome of this study demonstrated that the advanced deep learning model ECG-Lens outperformed the traditional algorithms, with an accuracy rate of 80% achieved by the ECG-Lens, Complex Convolutional Neural Network (CNN), which yielded the highest accuracy among all other models. Furthermore, the utilization of data augmentation techniques, such as the Stationary Wavelet Transform (SWT) algorithm, enhanced the results of the model. The conclusions drawn from this study provide valuable insights for healthcare providers and lay the groundwork for future research aimed at developing more targeted models for specific cardiac conditions.